%% file: compentition.tex
\documentclass[conference]{IEEEtran}
\usepackage{graphicx}
\usepackage{bbding}
\usepackage{overpic}
\usepackage{pgfplots}
\usepackage{overpic}
\usepackage{cite}

\usepackage[colorlinks,linkcolor=red, citecolor=green]{hyperref}

\pgfplotsset{compat=1.14}
\begin{document}

\title{\LARGE The Winning Solution to the iFLYTEK Challenge 2021 Cultivated Land Extraction from High-Resolution Remote Sensing Image}

\author{Zhen Zhao$^{1,3}$, Yuqiu Liu$^{1,3}$, Gang Zhang$^{2}$, Liang Tang$^{1}$, Xiaolin Hu$^{2,3,4}$\\
	\normalsize $^{1}$School of Technology, Beijing Forestry University, Beijing 100091, China.\\
	\normalsize $^{2}$Department of Computer Science and Technology, State Key Laboratory of Intelligent Technology and Systems, \\Beijing National Research Center for Information Science and Technology, Tsinghua University, Beijing 100084, China.\\
	\normalsize $^{3}$Tsinghua Laboratory of Brain and Intelligence(THBI), Tsinghua University, Beijing 100084, China.\\
	\normalsize $^{4}$Chinese Institute for Brain Research (CIBR), Beijing 100010, China}

\maketitle

\begin{abstract}
Extracting cultivated land accurately from high-resolution remote images is a basic task for precision agriculture. This report introduces our solution to the {\itshape iFLYTEK challenge 2021 cultivated land extraction from high-resolution remote sensing image}. The challenge requires segmenting cultivated land objects in very high-resolution multispectral remote sensing images. We established a highly effective and efficient pipeline to solve this problem. We first divided the original images into small tiles and separately performed instance segmentation on each tile. We explored several instance segmentation algorithms that work well on natural images and developed a set of effective methods that are applicable to remote sensing images. Then we merged the prediction results of all small tiles into seamless, continuous segmentation results through our proposed overlap-tile fusion strategy. We achieved the first place among 486 teams in the challenge.
\end{abstract}
\IEEEoverridecommandlockouts

\IEEEpeerreviewmaketitle

\section{Introduction}
The application of digital agricultural services usually needs to accurately count the location and area of farmland owned by farmers. This information lays the foundation for farmland scientific management and agricultural credit loan later. Automatic extraction of field boundaries from satellite imagery can reduce the reliance on manual input of farmland information, which is time-consuming\cite{waldner2020extracting}. To advance the research on this task, iFLYTEK and ChangGuang Satellite jointly held the challenge of extracting cultivated land from high-resolution remote sensing images. (\url{http://challenge.xfyun.cn/topic/info?type=plot-extraction-2021})

Some works have explored the application of deep learning models on high-resolution remote sensing images. Many works\cite{waldner2020extracting,li2021deep,persello2019delineation,zheng2021high,waldner2021detect} are inspired by the U-net\cite{ronneberger2015u} and they focused on the semantic segmentation task. However, semantic segmentation models do not necessarily yield separated cultivated land, and additional postprocessing is required to define regions with closed contours and obtain individual instances\cite{waldner2021detect}. While end-to-end instance segmentation models can extract individual instances in one go, they have not been widely tested\cite{waldner2021detect}. {\itshape Potlapally et al.}\cite{potlapally2019instance} and {\itshape Zhang et al.}\cite{zhang2018deep} developed models based on Mask R-CNN for segmentation of arctic ice-wedges and cultivated land, respectively. However, the particularity of remote sensing images makes the accuracy of algorithms less than satisfactory, requiring the application of more powerful algorithms and feature extractors to improve the segmentation performance.

In this report, we validated the applicability of various instance segmentation algorithms in the task of cultivated land segmentation on remote sensing images. Although the current instance segmentation methods perform well on natural images, they are rarely applicable to remote sensing images. Objects appear at arbitrary locations in remote sensing images with uneven distribution. There are usually no clear boundaries between objects. we had explored a set of effective solutions from different aspects, including data preprocessing, model pre-training, segmentation algorithm, and results postprocessing. Then, we successfully built a highly effective and efficient pipeline to solve this problem.

Among the whole segmentation pipeline, a very critical part is results postprocessing, i.e. merging predicted results from all small tiles.
We must cut a high-resolution image into many tiles as the limited GPU memory to perform instance segmentation. Combining results from multiple overlapping tiles into seamless, continuous segmentation results is rarely mentioned, but it is vital in practical engineering. {\itshape Olaf et al.}\cite{ronneberger2015u} took the average value of predictions inside the overlap area as the final semantic segmentation results, but this way is not applicable to instance segmentation.
{\itshape François et al.}\cite{waldner2021detect} only retained the detection result with the lowest instance uncertainty in the overlapping area between tiles, but we observed that many incomplete instances with high uncertainty values were kept. Instead of using instance uncertainty as a criterion, we heuristically divided each tile into the {\itshape ignore area} and {\itshape target area}, and only kept the detection results inside the target area. We further utilize matrix operation to shorten the total time of strategy significantly.

We explored an efficient pipeline and a more superior overlap-tile fusion strategy, achieving an $\rm AP_{50}$ score of 58.1 on the test set. We scored 63.38 points at the evaluation metrics of the iFLYTEK challenge and ranked the first place among 486 teams. Code and dataset are available at:

\url{https://github.com/zhaozhen2333/iFLYTEK2021.git}

\begin{figure*}
\centering
\begin{minipage}[t]{0.48\textwidth}
\centering
\includegraphics[width=1\textwidth]{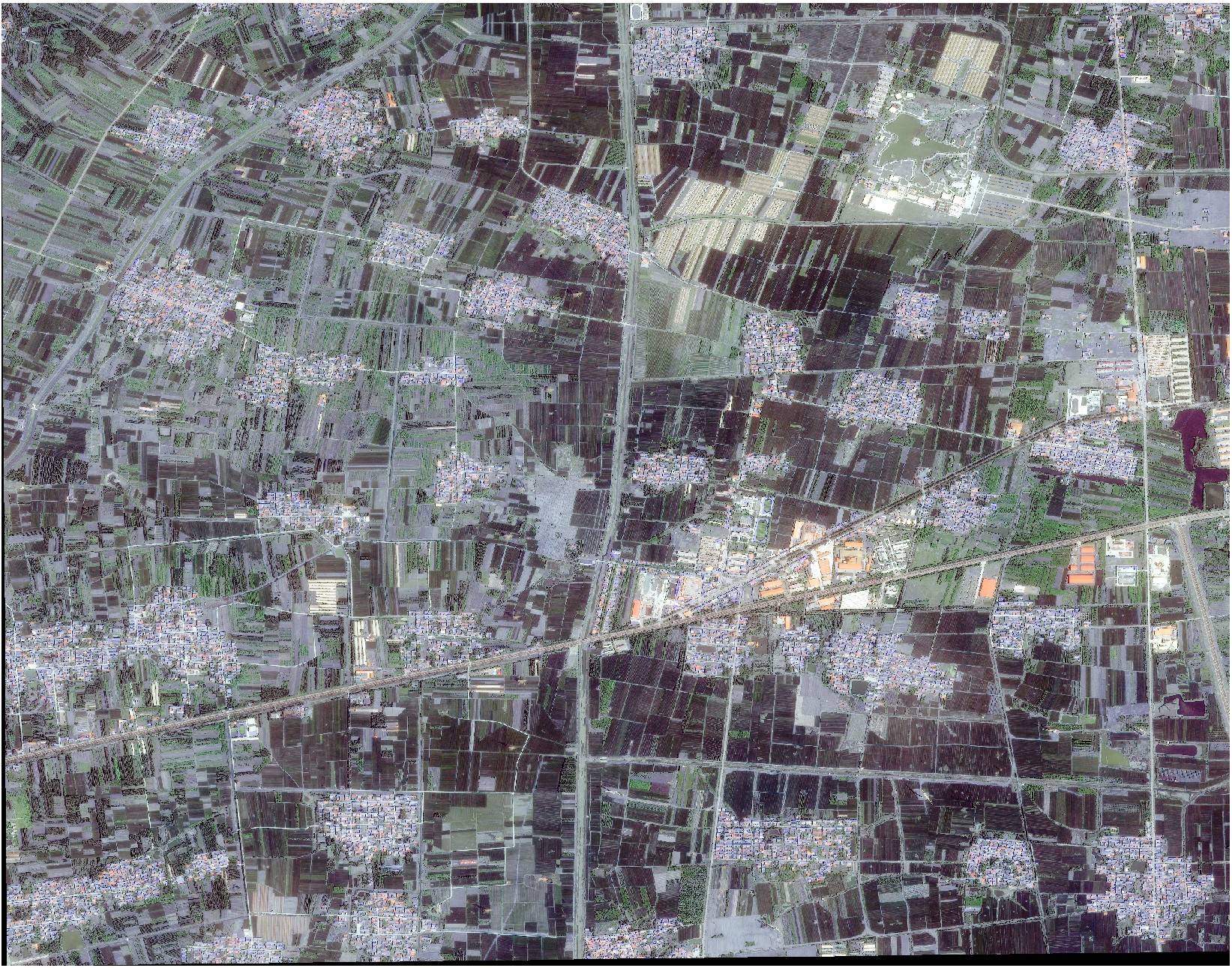}
\end{minipage}
\begin{minipage}[t]{0.48\textwidth}
\centering
\includegraphics[width=1\textwidth]{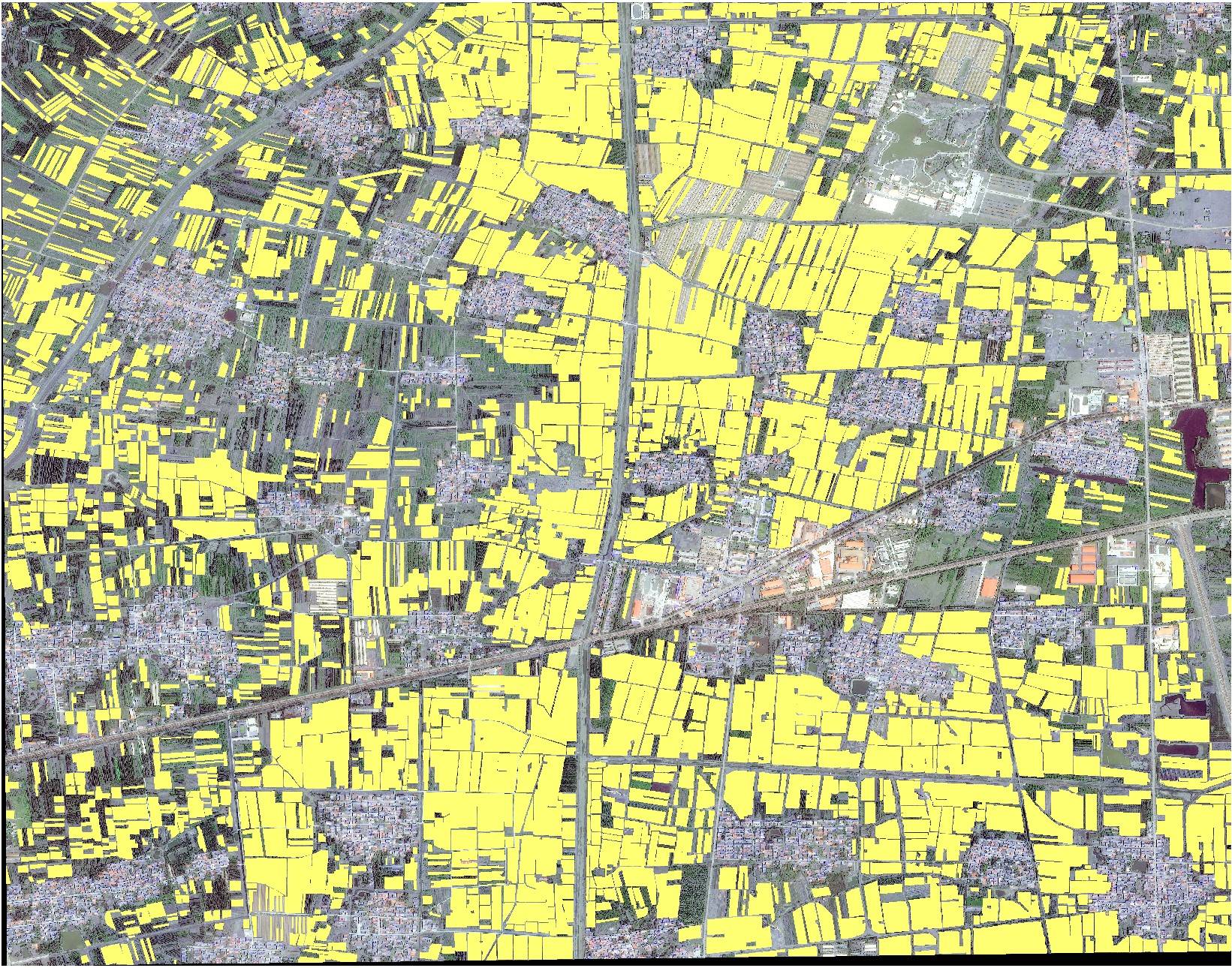}
\end{minipage}
\caption{ Examples of the JiLin-1 image dataset. Raw RGB image on the left; overlay with ground truth segmentation on the right.}
\end{figure*}

\section{Materials and Methods}
\subsection{ JiLin-1 image dataset}
The JiLin-1 image dataset was used in the 2021 {\itshape iFLYTEK challenge} of cultivated land extraction from high-resolution remote sensing images (Fig.1.). Each image has four bands: blue, green, red, and near-infrared. The spatial resolution of the JiLin-1 multispectral images is 0.75-1.1 M. However, their sizes are not fixed, and the pixels are between 5000-18000. The competition was divided into three rounds. Fifteen raw data was provided as the test set in the preliminary, and 18 raw data was provided in the second round. However, only 16 high-resolution images were supplied as training set in the preliminary.

\subsection{ Evaluation Criterion}
Calculation formula of preliminary and semi-final tests results is: \[ \rm Score1 = 0.6AP_{50} + 0.4mIoU \]
where, $\rm AP_{50}$ is the AP for instance segmentation at $\rm IoU > 0.5$, and $\rm mIoU$ is the average value of $\rm IoU$ for semantic segmentation over areas of cultivated land and background.
But we only regarded it as an instance segmentation task, ignoring the impact of semantic segmentation indicators $\rm mIoU$. More accurate instance segmentation results also correspond to more accurate semantic segmentation results.

In the final, the competition organizers would calculate the overall score of the algorithm, and the calculation formula is: $$ \rm Score2 = 0.5Score1 + 0.3Score_{eff} + 0.1Score_{cod} + 0.1Score_{doc} $$ where, $\rm Score_{eff}$ is the efficiency score of the algorithm, $\rm Score_{cod}$ is the evaluation score of the code, and $\rm Score_{doc}$ is the evaluation score of the technical documentation. Since both $\rm Score1$ and $\rm Score_{eff}$ had a high proportion, we tried to balance the performance and the running time of the whole pipeline. The competition organizers scored $\rm Score_{cod}$ and $\rm Score_{doc}$, and no specific rules had been announced.

\subsection{ Designs about the Whole Pipeline}

Our algorithm used the Hybrid Task Cascade(HTC)\cite{chen2019hybrid} model with a backbone of ResNeXt-101-64x4d\cite{xie2017aggregated} and Deformable ConvNets v2 (DCN)\cite{zhu2019deformable}. We used the model pre-trained for 20 epochs on the COCO dataset\cite{lin2014microsoft} for weight initialization, which was better than using the pre-trained model on ImageNet\cite{deng2009imagenet}. Finally, the results of the algorithm output were post-processed by the overlap-tile fusion strategy. Soft-NMS\cite{bodla2017soft} was used to improve the final performance. We also tried Content-Aware ReAssembly of FEatures (CARAFE)\cite{wang2019carafe}, DetectoRS\cite{qiao2021detectors}, PA-FPN\cite{liu2018path} to enhance the backbone of the model, but they did not work well. The multi-scales test did not achieve positive results. Random cropping and InstaBoost\cite{fang2019instaboost} for data augmentation also were useless. This may result from the fact that remote sensing images had simpler characteristics than the complicated scenarios and objects in COCO, and most of the cultivated land was densely distributed.

\subsection{ Overlap-Tile Fusion Strategy}

We used a sliding window of size 1536×1536, with a moving stride size of 1280, to cut out many tiles of pixels from the original images from top to bottom and left to right. We used heuristic rules to automatically divide each tile into {\itshape target area} T and {\itshape ignore area} (Fig.2.).
The left edge and the upper edge of the {\itshape ignore area} were 2 pixels away from the tile border respectively.
If and only if coordinates (x, y) of the upper left corner of the detection boxes fell within the target area, detection results were retained. When a tile was on the border of the original image, the {\itshape ignore area} that lay on edge in the same direction of the tile was merged into the {\itshape target area}. The position coordinates of tiles had been collected when making the dataset. 

We achieved 5.8 points $\rm Score1$ improvements over direct outputs. We also used matrix operation to shorten the total inference time significantly. In fact, the matrix merging operation takes almost no time. 

\begin{figure}
\centering
\includegraphics[width=0.48\textwidth]{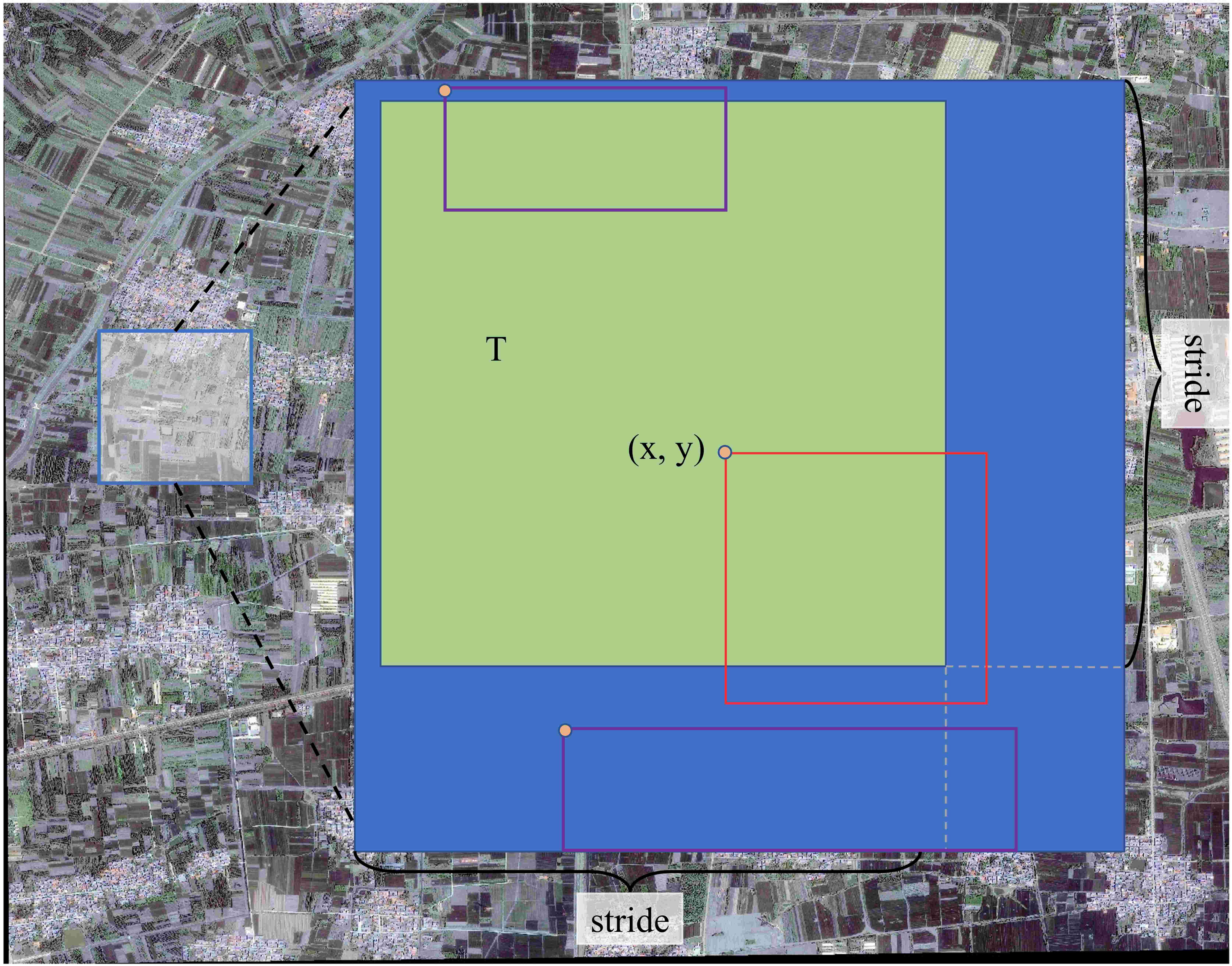}
\caption{ Overlap-tile fusion strategy. The detection result is preserved if and only if the upper left corner of the detected bounding box (x, y) lies in the target area T}
\end{figure}

\section{Experiments}
We trained our model on eight 2080Ti GPUs. Our code was developed based on the mmdetection\cite{chen2019mmdetection} codebase. Unless otherwise specified, all parameters followed their default settings.
\subsection{ Data Preprocessing}
We used the sliding window to divide the 16 raw data of the original training set into tiles of pixels from top to bottom and left to right. The window size was 512×512, the stride of window movement was also 512. Then, we divided these tiles into a training set and a validation set with a ratio of 5:1. The training and the validation set include 3115 tiles and 629 tiles, respectively. The size of tiles on the lower and left edges of raw data is smaller than 512.

To obtain the seamless segmentation results for the test set, we used the same sliding window method to create a new test set. The window size was 1536×1536, and the step of window movement was 1280. It would help us apply our overlap-tile fusion strategy.

The processed training set contains a large number of images without cultivated land. We let these images participate in the training process, which could slightly improve the performance of our model.

We also found that the natural images of R, G and B bands achieved better performance than the combination of NIR, G and B bands. The combination of NIR, G and B bands, is commonly used in remote sensing to enhance the vegetation characteristics of the surface.
\subsection{ Model-Related Experiments}

\subsubsection{ Model}
We did ablation experiments on different models. Table \uppercase\expandafter{\romannumeral1} shows the effect comparison between the HTC and the classic Mask R-CNN\cite{he2017mask}, and both models use ResNet-50 as backbone. The results shows that HTC is a more powerful detection model, as it effectively improves the information flow, not only across stages but also between tasks. We took the better HTC model as our baseline.
\begin{table}
\renewcommand{\arraystretch}{1.3}
\caption{ Effects of models.}
\begin{center}
\begin{tabular}{c c c}
\hline
Model & $\rm AP_{val}$ & $\rm AP_{val50}$\\
\hline
Mask R-CNN & 41.9 & 68.6\\
{\bfseries HTC} & {\bfseries 45.2} & {\bfseries 70.6}\\
\hline
\end{tabular}
\end{center}
\end{table}

\subsubsection{ Techniques for Enhancing Backbone}
We first did ablation experiments on different backbones. Table \uppercase\expandafter{\romannumeral2} shows the impact of different backbones on the performance of Mask R-CNN. We chose the best ResNeXt-101 as the base backbone. We noticed that the HTC model needed longer time to converge, so we also conducted ablation experiments on the training time. The results are shown in Table \uppercase\expandafter{\romannumeral3}. The evaluation metric for instance segmentation in the competition was the $\rm AP_{50}$, and the $\rm AP_{50}$ of the HTC model had converged within 12 epochs. 
Then, to further improve the performance of the model, we studied some techniques that have been proved to be effective in general object detection to improve the backbone of the model (Fig.3.). Techniques such as DCN significantly enhanced the performance. CARAFE, DetectoRS, PA-FPN had no significant changes or adverse effects, although they performed well on the COCO dataset. It shows the particularity of the remote sensing image dataset, perhaps because it has simpler characteristics than the complicated scenarios and objects in COCO. In the following experiments, we finally used ResNeXt-101-64x4d+DCN as backbone.
In the final experiment, we observed that the natural image composed of R, G and B might show the cultivated land objects more clearly than the combination of NIR, R and G. Therefore, a new data format was used here, so the AP was increased by 1.5\%. Except for replacing the band combination, all other settings were the same as the original dataset.

\begin{table}
\renewcommand{\arraystretch}{1.3}
\caption{ Effects of backbones.}
\begin{center}
\begin{tabular}{c c c}
\hline
Backbones & $\rm AP_{val}$ & $\rm AP_{val50}$\\
\hline
ResNet-50\cite{he2016deep} & 40.0 & 65.9\\
ResNet-101 & 41.1 & 67.1\\
{\bfseries ResNeXt-101} & {\bfseries 41.4} & {\bfseries 67.8}\\
ResNeSt-101\cite{zhang2020resnest} & 37.3 & 62.2\\
\hline
\end{tabular}
\end{center}
\end{table}

\begin{table}
\renewcommand{\arraystretch}{1.3}
\caption{ Effects of epochs and cardinalities.}
\begin{center}
\begin{tabular}{c c c c}
\hline
Backbones & Epochs & $\rm AP_{val}$ & $\rm AP_{val50}$\\
\hline
X101-64x4d & 12 & 46.7 & 71.9\\
{\bfseries X101-64x4d} & {\bfseries 20} & {\bfseries 47.3} & {\bfseries 72.1}\\
X101-64x4d & 24 & 47.2 & 72.0\\
X101-64x4d & 36 & 46.7 & 71.4\\
\hline
\end{tabular}
\end{center}
\end{table}

\begin{figure}
\centering
\includegraphics[width=0.48\textwidth]{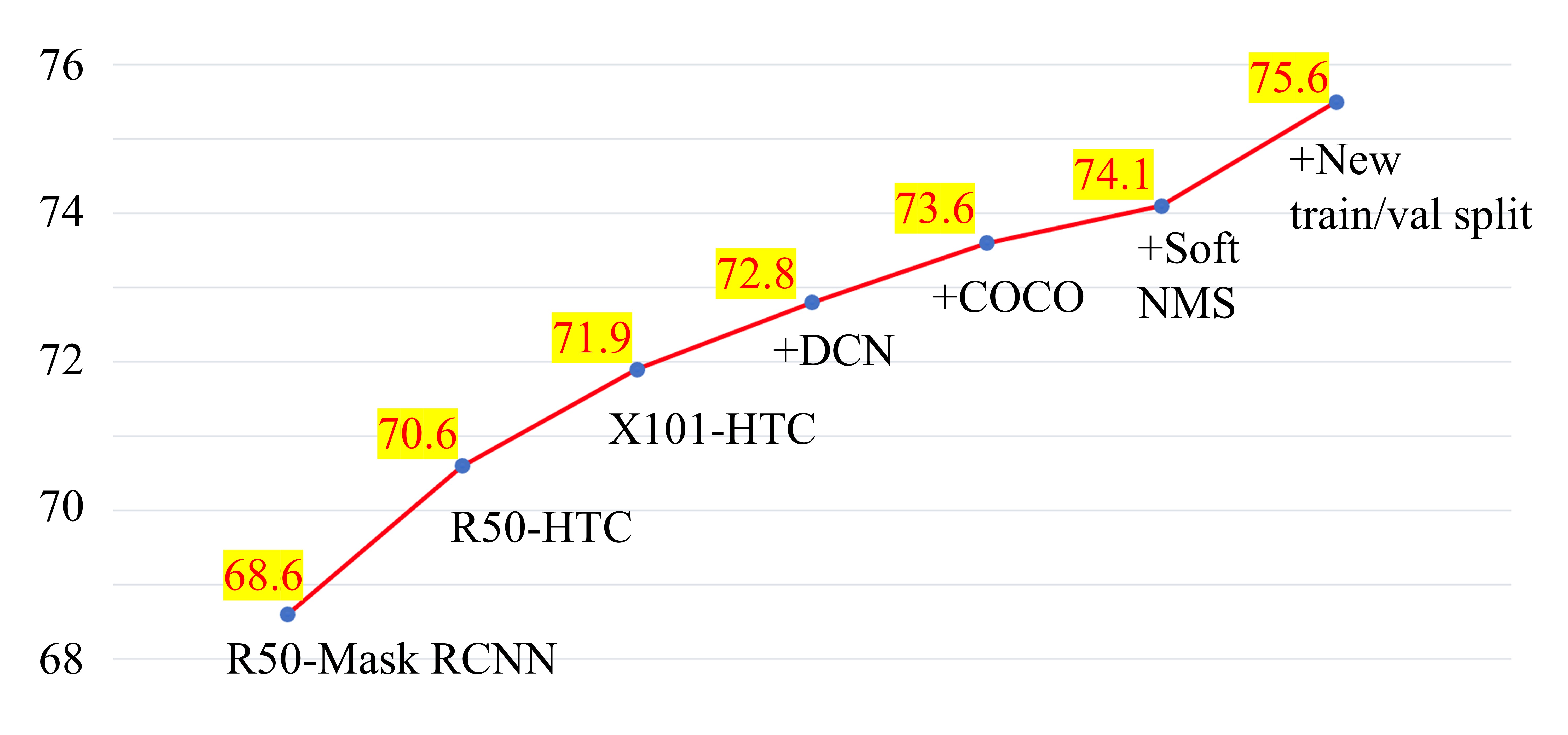}
\caption{ Effects of model pre-training and techniques for enhancing backbone.}
\end{figure}

\subsubsection{ Model Pre-training}
Using pre-trained models can reduce the time for model convergence and usually achieve better results. HTC with ResNeXt-101-64x4d+DCN backbone trained for 12 epochs was chosen as our baseline for ablation studies. Using the COCO pre-trained model can achieve better performance on remote sensing images than the ImageNet pre-trained model. The results are shown in Table \uppercase\expandafter{\romannumeral4}.

\begin{table}
\renewcommand{\arraystretch}{1.3}
\caption{ Effects of pre-trained models.}
\begin{center}
\begin{tabular}{c c c}
\hline
Pre-trained models & $\rm AP_{val}$ & $\rm AP_{val50}$\\
\hline
ImageNet & 47.5 & 72.8\\
{\bfseries COCO} & {\bfseries 49.6} & {\bfseries 73.6}\\
\hline
\end{tabular}
\end{center}
\end{table}

\subsection{ Overlap-tile Fusion Strategy}
As described in Sect.\uppercase\expandafter{\romannumeral2}.D, We proposed that the robust overlap-tile fusion strategy should meet the following rules (Fig.2.):  1) The tiles we got should be as large as possible to ensure that each cultivated land object would appear entirely in at least one of the tiles.
2) If and only if coordinates (x, y) of the upper left corner of the detection boxes fell within the target area, detection results were retained. 
3) When a tile was on the border of the original images, the {\itshape ignore area} that on edge in the same direction of the tile was merged into the {\itshape target area}.

In order to merge the mask results with less performance loss, the stride should be smaller, such as 256, but this would increase the test time. Considering that, the size of the test set we generated was 1536, and the moving stride size was 1280. The proportion of large objects in the dataset was low, so our approach lost slightly.
One of the mask predictions output by the overlap-tile fusion strategy is shown in Figure 4.
\begin{figure}
\centering
\includegraphics[width=0.48\textwidth]{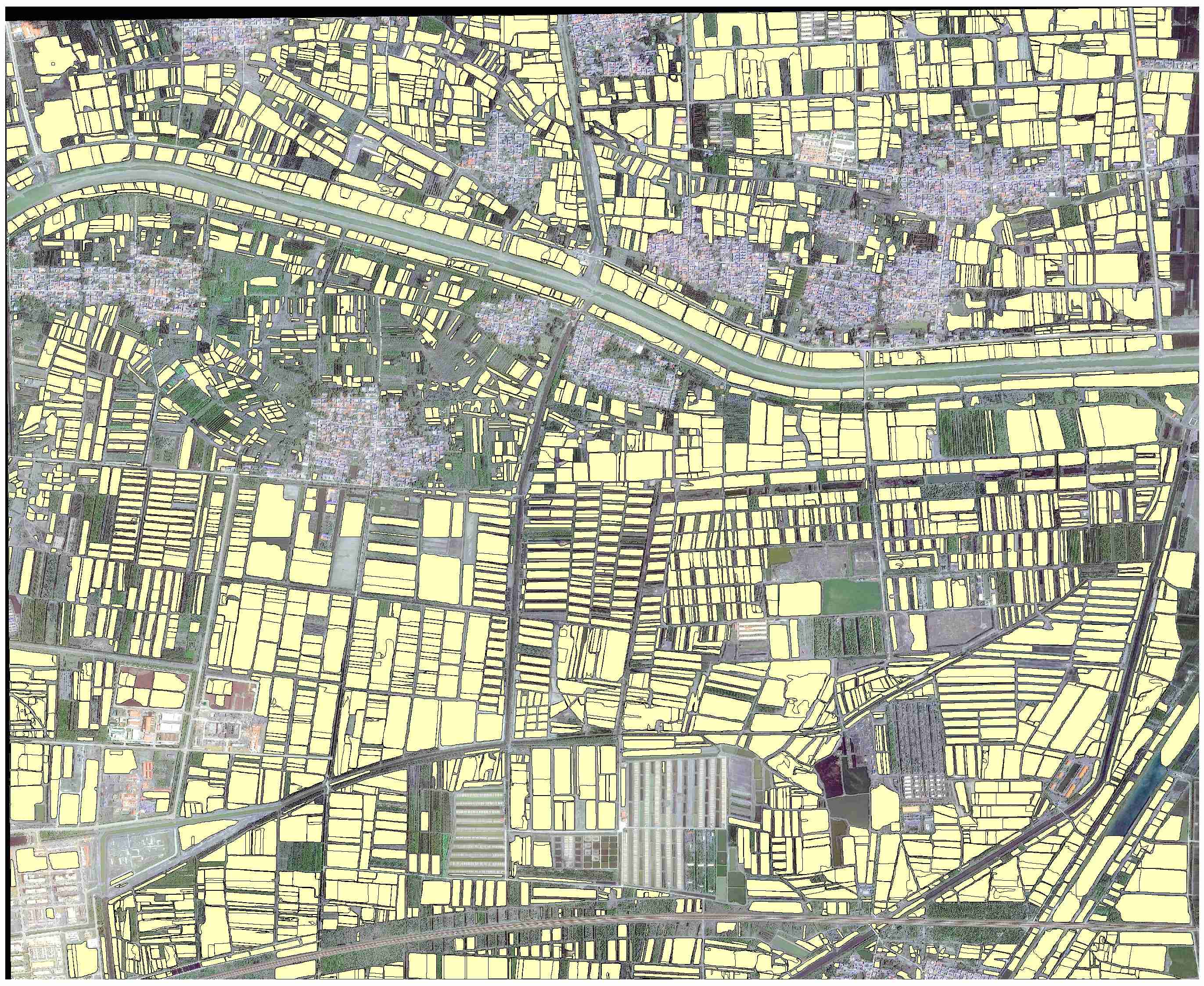}
\caption{ Complete output results.}
\end{figure}

\subsection{ FINAL RESULTS}
The whole process of $\rm AP_{50}$ growth can be seen intuitively in Figure 3. Including splitting raw data, model processing, and fusing prediction results, and the entire test process of 18 raw data only took 24 minutes on eight 2080Ti GPUs. If we adjusted the image scale for model training and sacrificed the value of Score1 slightly, the efficiency of the model would be significantly improved. Results were shown in Table \uppercase\expandafter{\romannumeral5}. The Score1 of the top five teams in the competition are shown in Figure 5. We are the first team. The results in Figure 5 are not expressed as percentages.

\begin{table}
\renewcommand{\arraystretch}{1.3}
\caption{ Efficiency of our models.}
\begin{center}
\begin{tabular}{c c c c}
\hline
Training scales & Time/min & $\rm AP_{test50}$ & Score1\\
\hline
600 & 15 & 55.7 & 62.88\\
800 & 24 & 58.1 & 63.38\\
\hline
\end{tabular}
\end{center}
\end{table}

\begin{figure}
\centering
\includegraphics[width=0.48\textwidth]{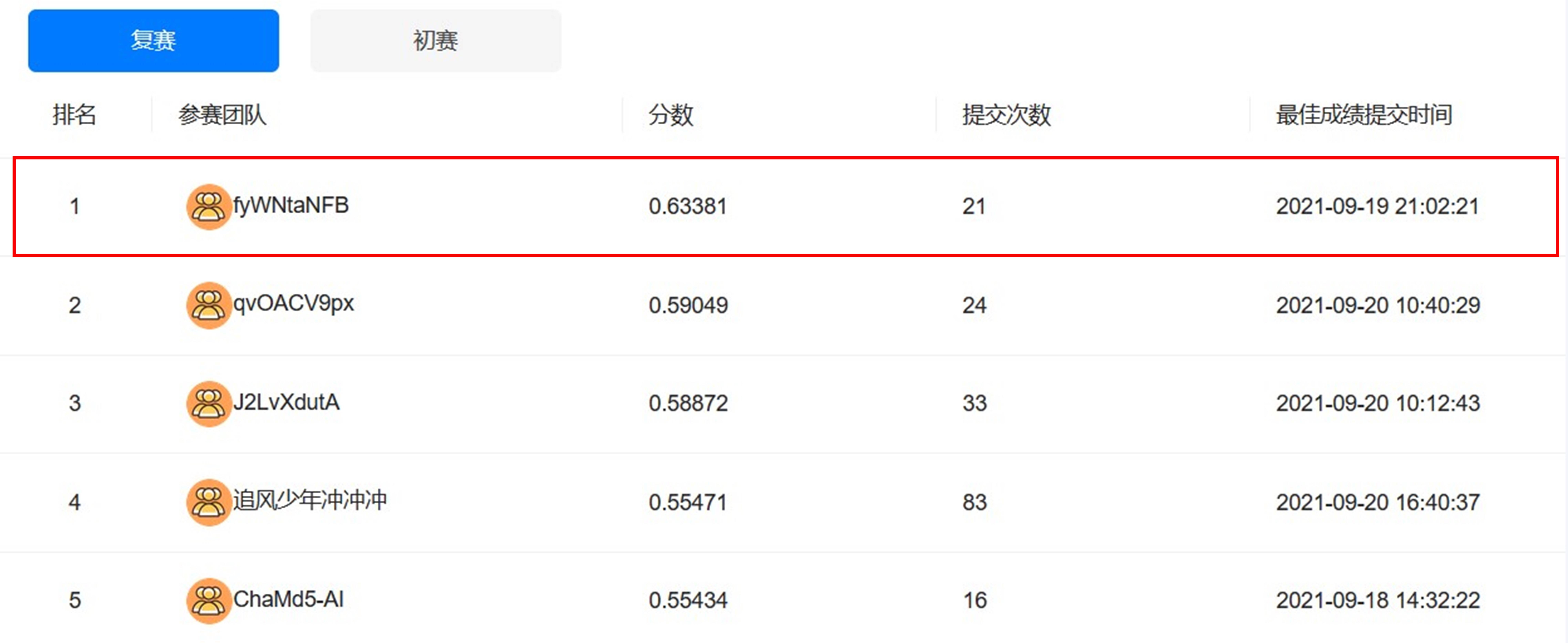}
\caption{ The screenshot of the final results announced on the competition website. The top five teams are shown.}
\end{figure}

\section{ CONCLUSION}
In this work, we successfully built a highly effective and efficient pipeline for cultivated land extraction in high-resolution remote sensing images. We found that the commonly used technologies that work well on the COCO dataset do not perform well on remote sensing images. For large images that GPU cannot deal with, we provided a more detailed overlap-tile fusion strategy to generate seamless and continuous segmentation results. On the dataset of this competition, Score1 of our pipeline reached 63.38, while the total duration was only 24 minutes. We ranked the first place among 486 teams. We hope that our work can promote the application of digital agriculture in rural areas.

\section*{ Acknowledgement}
This work was supported by the National Natural Science Foundation of China (Nos. U19B2034 and 62061136001) and THU-Bosch JCML center.

\input{compentition.bbl}

\bibliographystyle{IEEEtran}

\end{document}